\documentclass[letterpaper, 10 pt, conference]{ieeeconf}  %

\IEEEoverridecommandlockouts                              %

\usepackage{algorithm}
\usepackage{algorithmic}
\usepackage{float}
\usepackage{bbm}
\usepackage{graphicx}
\usepackage{subcaption}
\usepackage{xcolor}
\usepackage{hyperref}       %
\usepackage{url}            %
\usepackage{booktabs}       %
\usepackage{amsfonts}       %
\usepackage{amsmath,amssymb}
\usepackage{nicefrac}       %
\usepackage{wrapfig}
\usepackage{multirow}
\usepackage{pifont}
\usepackage{booktabs}
\usepackage{bbold}
\usepackage{mathtools}

\newcommand{\Skip}[1]{}

\newcommand{\ie}{i.e.\ }

\DeclareMathOperator*{\argmin}{arg\,min}

\title{\LARGE \bf
Efficient Skill Acquisition for Complex Manipulation Tasks \\ in Obstructed Environments

\author{Jun Yamada$^{1}$, %
Jack Collins$^{1}$,
Ingmar Posner$^{1}$ %
}

\thanks{$^{1}$Applied AI Lab (A2I), Oxford Robotics Institute, University of Oxford}%
\thanks{Correspondence to: {\tt\small jyamada@robots.ox.ac.uk}}%
}

\renewcommand{\baselinestretch}{0.975}

\begin{document}

\maketitle
\thispagestyle{empty}
\pagestyle{empty}

\begin{abstract}

Data efficiency in robotic skill acquisition is crucial for operating robots in varied small-batch assembly settings. 
To operate in such environments, robots must have robust obstacle avoidance and versatile goal conditioning acquired from only a few simple demonstrations.
Existing approaches, however, fall short of these requirements.
Deep reinforcement learning (RL) enables a robot to learn complex manipulation tasks but is often limited to small task spaces in the real world due to sample inefficiency and safety concerns.  
Motion planning (MP) can generate collision-free paths in obstructed environments, but cannot solve complex manipulation tasks and requires goal states often specified by a user or object-specific pose estimator.
In this work, we propose a system for efficient skill acquisition that leverages an object-centric generative model (OCGM) for versatile goal identification to specify a goal for MP combined with RL to solve complex manipulation tasks in obstructed environments.
Specifically, OCGM enables one-shot target object identification and re-identification in new scenes, allowing MP to guide the robot to the target object while avoiding obstacles.
This is combined with a skill transition network, which bridges the gap between terminal states of MP and feasible start states of a sample-efficient RL policy.
The experiments demonstrate that our OCGM-based one-shot goal identification provides competitive accuracy to other baseline approaches and that our modular framework outperforms competitive baselines, including a state-of-the-art RL algorithm, by a significant margin for complex manipulation tasks in obstructed environments.  

\end{abstract}

\section{Introduction}
Teaching new skills to robots using limited supervision is essential for maximising the up-time and productivity of robots, leading to faster return on investment. Small-batch manufacturing, where there are a limited number of parts to be produced, is an exemplary environment that would greatly benefit from efficient skill acquisition. In a small-batch setting, a robot must learn to manipulate new objects while maintaining data efficiency in potentially arbitrarily obstructed environments.
However, existing methods such as motion planning and reinforcement learning struggle to satisfy such requirements.

Motion planning (MP)~\cite{amato1996PRM, lavalle1998rapidly, karaman2011sampling} generates collision-free paths capable of guiding a robot safely in obstructed environments given an explicit state of the environment and goal.
However, MP is not designed to plan through complex manipulation tasks requiring environmental interaction.
Furthermore, MP necessitates the specification of a goal state in the robot's frame of reference, which is typically accomplished through manual engineering~\cite{khodeir2021learning}, template matching~\cite{le2019matching}, or an object-specific pose estimator~\cite{lee2020guided} trained on manually labeled data.

Deep reinforcement learning (RL), on the other hand, has shown promising outcomes in learning to control a robot for complex manipulation tasks such as grasping~\cite{kalashnikov2018qtopt, zhan2020framework} and insertion~\cite{luo2021shield}.
However, prior works often limit operation to simulated
environments~\cite{haarnoja2018sac} or heavily restrict and regulate operating spaces by executing with a short horizon without obstructions~\cite{luo2021shield, zhan2020framework} due to the sample inefficiency and potential of executing unsafe policies.

\begin{figure}[t]
    \centering
    \includegraphics[width=0.6\linewidth]{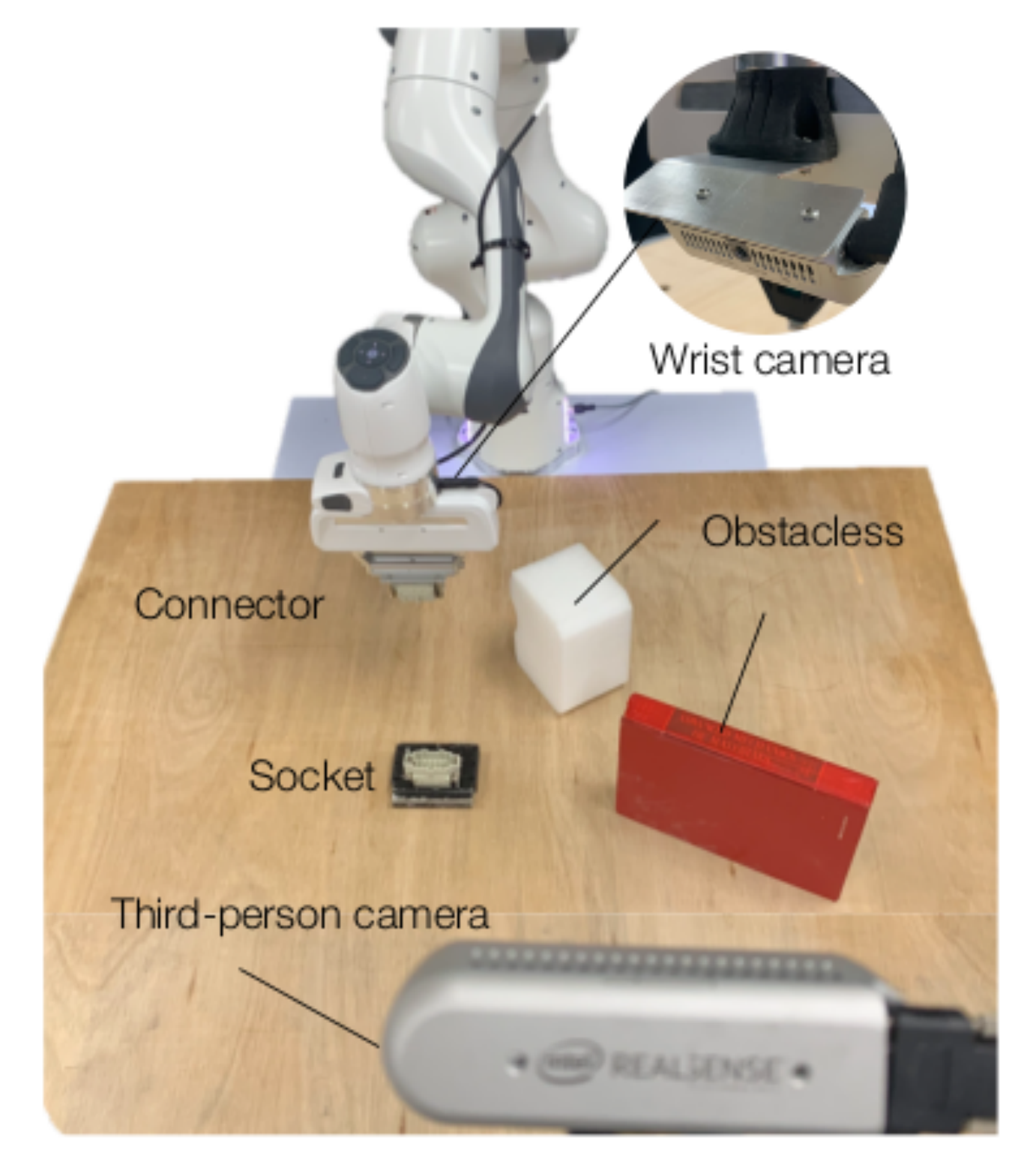}
    \caption{
        \textbf{Task setup.} We solve complex manipulation tasks within the entire operational space of a robot by leveraging an OCGM for versatile and efficient goal acquisition paired with MP and RL. Note that obstacles and a socket are randomly placed on the table.
    }
    \label{fig:task_setup}
    \vspace{-1.3em}
\end{figure}

Combining MP and RL has been investigated by
several prior works~\cite{yamada2020mopa, lee2020guided} and shows the potential of leveraging the strengths of both methods to solve manipulation tasks in obstructed environments.
Yet, goal specification for MP in prior work has relied on either sample-inefficient interaction with the environment or an object-specific pose estimator, which needs re-training for each new target object.

In this work, we take advantage of both MP and RL to achieve \textit{efficient} skill acquisition for complex manipulation tasks in environments with arbitrary obstacles.
We propose a system that leverages an object-centric generative model (OCGM)~\cite{wu2021Apex} for versatile, one-shot goal identification and re-identification as a key component in solving complex manipulation tasks from a few simple human demonstrations.
Specifically, we identify a target object from a \textit{single} demonstration using an OCGM, pre-trained on diverse synthetic scenes, leading to robust re-identification of that object in new scenes by matching to its object-centric representation.
Using the object's position as a goal, the motion planner generates a collision-free path to the target object while avoiding obstacles before a learned RL policy is executed to complete the complex manipulation tasks.
We train an RL policy for each manipulation skill from a sparse reward to preclude specialist knowledge for reward engineering and guide exploration using a handful of easy-to-collect demonstrations.
To maximise performance, we also introduce a skill transition network to reduce failures that occur when transitioning from MP to the learned RL policy.

The contributions of our work are fourfold: (1) we propose a system for efficient skill acquisition in obstructed environments that leverages an OCGM for object-agnostic, \textit{one-shot} goal specification, (2) we introduce a transition network that smoothly interpolates between terminal states of motion planning and feasible start states of a learned RL policy, 
(3) we show that our OCGM-based one-shot goal specification method achieves comparable accuracy against several goal identification baselines, and (4) we demonstrate that our system performs significantly better in real-world environments compared to baselines, including a state-of-the-art RL algorithm.

\section{Related Works}
Recent success in deep RL~\cite{kalashnikov2018qtopt, haarnoja2018sac, gu2016deep, levine2015endtoend} enables a robot to learn complex manipulation tasks such as grasping~\cite{kalashnikov2018qtopt, zhan2020framework} and insertion~\cite{luo2021shield, vecerik2018practical, lee2018making, Davchev_2022, carvalho2022residual} driven by a reward.
To avoid the requirement of specialist knowledge for reward engineering, several prior works have proposed sample-efficient RL methods that can learn complex manipulation skills from a sparse reward by leveraging a small number of demonstrations for guided exploration~\cite{zhan2020framework, luo2021shield, vecerik2017leveraging, vecerik2018practical}. 
However, due to the sample inefficiency of sparse rewards, studies have been primarily conducted in simulated environments or within limited task spaces in the real world.
Learning from demonstration (LfD)~\cite{SCHAAL1999233, billard2008, groth2021vmc} is an alternative method for a robot to learn manipulation tasks by imitating behaviour in expert demonstrations collected by a human operator, but it often requires a large number of demonstrations to acquire manipulation skills.
While InsertionNet~\cite{spector2021insertion} enables a robot to solve industrial insertion tasks within the entire operational space of a robot manipulator from a small number of demonstrations, it is evaluated in a clean environment without obstruction. Successful insertion is also made possible by a small initiation set for the learnt skill.
Adaptive LfD for insertion has also been proposed \cite{wen2022you}, allowing a policy to quickly adapt to a new industrial object from the same category seen in training using only a single demonstration and the object mesh.
However, such mesh information is not readily available, limiting real-world applications.
In our work, a manipulation skill is learnt using Framework for Efficient Robot Learning (FERM)\cite{zhan2020framework} but is essentially interchangeable with any method capable of learning the needed skill efficiently.

Motion planning (MP)~\cite{amato1996PRM, kavraki1994randomized, lavalle1998rapidly, lavalle2000rapidly, karaman2011sampling} can effectively generate a collision-free path from a robot's initial configuration to a goal pose using an explicit model of the robot and environment. 
However, such a goal pose is often specified by a user or object-specific pose estimator to generate a path.
Further, MP does not model the dynamics of the surrounding environment or objects and therefore complex manipulation tasks are out of the scope for MP.

Several previous works~\cite{yamada2020mopa, lee2020guided, xiali2020relmogen, kuo2021uncertainty, scholz2010combining} combine MP and RL to leverage the benefits of both methods to solve manipulation tasks in unstructured environments. However, these preceding works limit their real-world applicability by requiring a large number of samples to learn a goal estimator \cite{yamada2020mopa} or by retraining an object-specific predictor for each new goal object \cite{lee2020guided}.

Our work leverages OCGMs~\cite{wu2021Apex, locatello2020slot, Lin2020SPACE} to find a target object for MP, negating the need for object-specific goal estimators. OCGMs learn structured representations of objects within complex scenes and provide object-specific encodings useful for matching, however, most OCGMs have never been applied to real world environments.
In contrast to goal specification methods that require human intervention or a large, object-specific datasets, i.e. template matching or object classifier, OCGMs hold the promise of versatile target object identification.
This work leverages APEX~\cite{wu2021Apex} by training the model in an unsupervised manner on a wide distribution of simulated data to assist with generalisation to real-world environments.

\section{Methodology}
\label{sec:method}
\begin{figure*}[t!]
\centering
\includegraphics[width=0.9\textwidth]{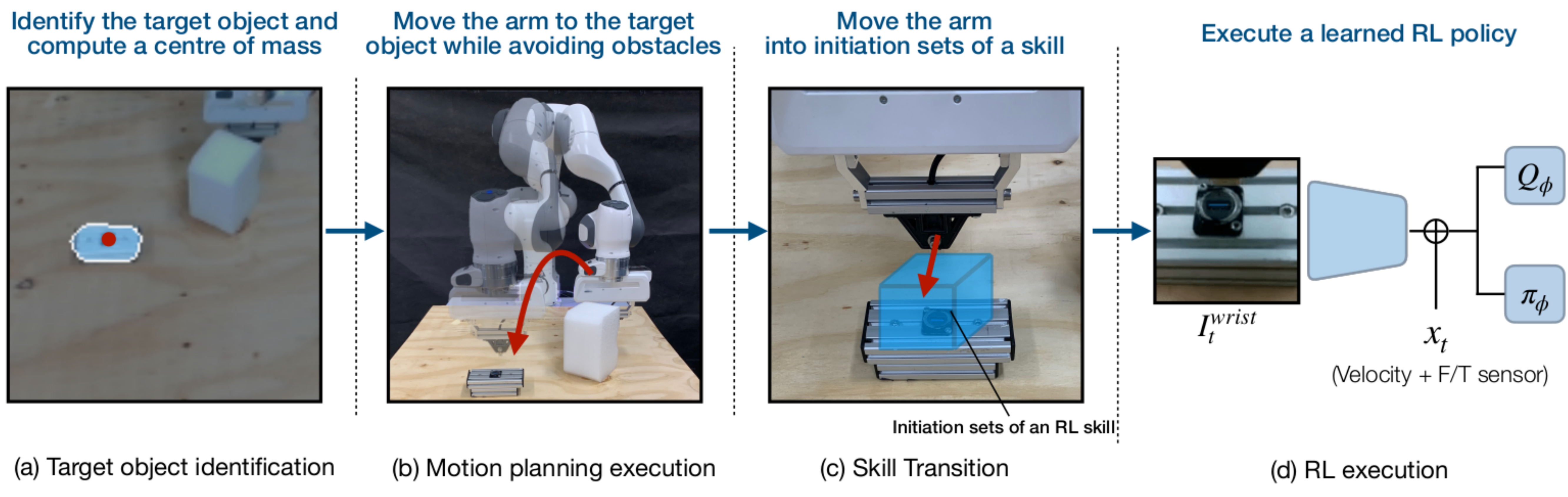}
\caption{\textbf{Our framework architecture.} (a) We leverage an OCGM to re-identify a target object such that its object-centric representation matches one extracted from a single demonstration. The goal state is specified in the robot's reference frame using an external RGB-D camera with calibrated extrinsics. (b) Given the goal state acquired in (a), a motion planner generates a collision-free path to the goal. (c) A skill transition network guides the arm from the terminal state of the motion planning (MP) to the initiation set of the RL policy. (d) Given a wrist camera image $\boldsymbol{I}^{wrist}_{t}$ and robot's internal state $\boldsymbol{x}_{t}$, a learned RL policy executes the final interaction until task completion.}
\label{fig:method}
\vspace{-1em}
\end{figure*} 

In this work, we present an efficient solution for solving complex manipulation tasks in obstructed, real-world environments by leveraging an OCGM.
We demonstrate our method on several industrial insertion tasks as they require learning complex insertion skills and also require the identification of the target socket to complete the tasks (see Figure~\ref{fig:task_setup} for our task setup).
We break our method down into pre-training, in-situ skill training and execution in the following subsections. The pre-training component of our method is only completed \textit{once} and can be reused for all future insertion tasks. In-situ skill training is required for each new insertion task and execution describes the process for autonomous task execution after training.

\subsection{Pre-training}

Pre-training is required for APEX~\cite{wu2021Apex}, our choice of unsupervised OCGM, to achieve versatile one-shot target object acquisition, but it only needs to be done once as it is trained on a diverse dataset to encourage generalise to a variety of real-world objects.
APEX is formulated as a set of VAEs, and takes a video sequence $\boldsymbol{I}_{1:T}$ as input. Each frame is decomposed into a set of latent representations for each discovered object $j$ consisting of object location $\boldsymbol{z}^{where}_{t,j}$, appearance $\boldsymbol{z}^{what}_{t, j}$, and presence $z^{pre}_{t,j} \in [0, 1]$, where $T$ is the number of frames in a video sequence. 
We train APEX on a synthetic dataset consisting of a set of trajectories in which a robot interacts with a diverse set of primitive shapes of differing colour and size.
Domain randomisation~\cite{tobin2017domain} is a common technique for transferring models trained on synthetic data to the real world, however, APEX is unable to handle drastic background changes present in domain randomised datasets because a changing background is hard to distinguish from a moving foreground object.
Instead, we add a small amount of noise to the camera pose for each trajectory, leading to variations in the images.
As a result, APEX is successfully applied to real-world scenes with similar background textures.
See Appendix~\ref{appendix:apex} for further information regarding APEX.

\subsection{In-Situ Skill Training}
\label{sec:goal_spec}

This section details the task-specific data and training required by our method. The data must be collected for each new task that the robot is taught, however, the supervised component only requires about 10 minutes to collect.
 First, a single demonstration,  ${\mathcal{D}^{goal}=\{(\boldsymbol{I}^{ext}_{t}, \boldsymbol{x}^{ee}_{t}), \dots\}}$ consisting of a sequence of images $\boldsymbol{I}^{ext}$ from the third-person camera and robot end-effector positions $\boldsymbol{x}^{ee}_{t}$, of a successful task completion from anywhere within the robot's operational space is collected for goal specification using the pre-trained APEX. 
 Additionally, 25 demonstrations, $D^{RL}=\{(\boldsymbol{I}^{wrist}_{t}, \boldsymbol{x}_{t}, \boldsymbol{a}_{t}), \dots\}^{|25|}$, of the complex interaction with objects for efficient RL training, are collected from within a limited task space such that the connector is always within sight of the wrist camera, where $\boldsymbol{I}^{wrist}_{t}$, $\boldsymbol{x}_{t}$, $\boldsymbol{a}_{t}$ are a wrist camera image, robot states, and action at time step $t$.

We train the RL policy with FERM~\cite{zhan2020framework} to complete the insertion task, which takes between $60$ to $90$ minutes to train on a desktop computer with an i7 processor and a Nvidia Titan X GPU.
FERM is composed of Soft Actor Critic~\cite{haarnoja2018sac}, contrastive learning~\cite{wu2018unsupervised, laskin_srinivas2020curl}, and image augmentation~\cite{laskin2020reinforcement}.
In addition to a gray-scale image from the wrist-mounted camera ($\boldsymbol{I}^{wrist[H\times W]}$ where $H$ and $W$ are 64 pixels), the policy takes as input the end-effector Cartesian velocity and F/T sensor data (see Figure~\ref{fig:method} (d)) and outputs the desired 3-dimensional Cartesian end-effector velocity for the robot. Because the policy takes as input local information, it is able to generalise to any location in the robot's operational space.
The RL policy is trained using a sparse reward $r_{t} = \mathbb{1}[\boldsymbol{s} \in S_{g}]$ where $S_{g}$ is a set of goal states defined as a goal pose of the the robot with $1cm$ tolerance determined by simply teleoperating the robot to the goal pose before training the policy.
To accelerate training of the RL policy, we leverage the task demonstrations $D^{RL}$ to initialise a replay buffer for guided exploration similar to FERM and train the policy and critic asynchronously, as inspired by prior work~\cite{luo2021shield}. We also limit the task space for this stage, such that the socket is always within sight of the wrist camera for training, improving sample efficiency and reducing the chance of unsafe interactions. For further details, see Appendix~\ref{appendix:RL}.
While we adopt FERM to train a low-level manipulation policy, it is interchangeable with any approach, including imitation learning~\cite{SCHAAL1999233, billard2008}, that can learn complex manipulation skills efficiently from a small number of demonstrations. Note, however, that the learned policy often requires a large coverage of states around a target object such as a socket for insertion tasks to increase the success rate.

We also introduce a skill transition network, inspired by prior work~\cite{johns2021coarse-to-fine}, for each insertion task to improve the task success rate. The terminal state of the MP is not guaranteed to be within the feasible start states of the RL policy, defined as \emph{the initiation set of the skill} (represented as a blue box in Figure~\ref{fig:method} (c)), due to the accumulation of errors in camera extrinsics and estimation of the MP goal state 3D.
To mitigate this issue, a simple convolutional neural network (CNN) is trained on data collected in a self-supervised manner to predict the Cartesian offset required to move the end-effector from the terminal states of MP to the initiation set of the skill.
The dataset is collected in less than 30 minutes by sampling random Cartesian poses around the target object and recording the sequences of wrist camera images $\boldsymbol{I}^{wrist}_{t}$ and offsets between the current end-effector pose and an initial pose used for RL training. The collected data contains only local information conditioned on the wrist camera which allows the transition network to generalise to unseen target positions.
For further details, see Appendix~\ref{appendix:transition}.

\subsection{Execution}

The execution of the task can be completed from \emph{anywhere} within the robot's operational space to any goal location. Execution follows four steps (see Figure~\ref{fig:method}) that are completed autonomously: (i) goal identification via OCGM, (ii) MP, (iii) skill transition network, and (iv) RL policy.

\begin{figure}
\centering
\includegraphics[width=0.5\textwidth]{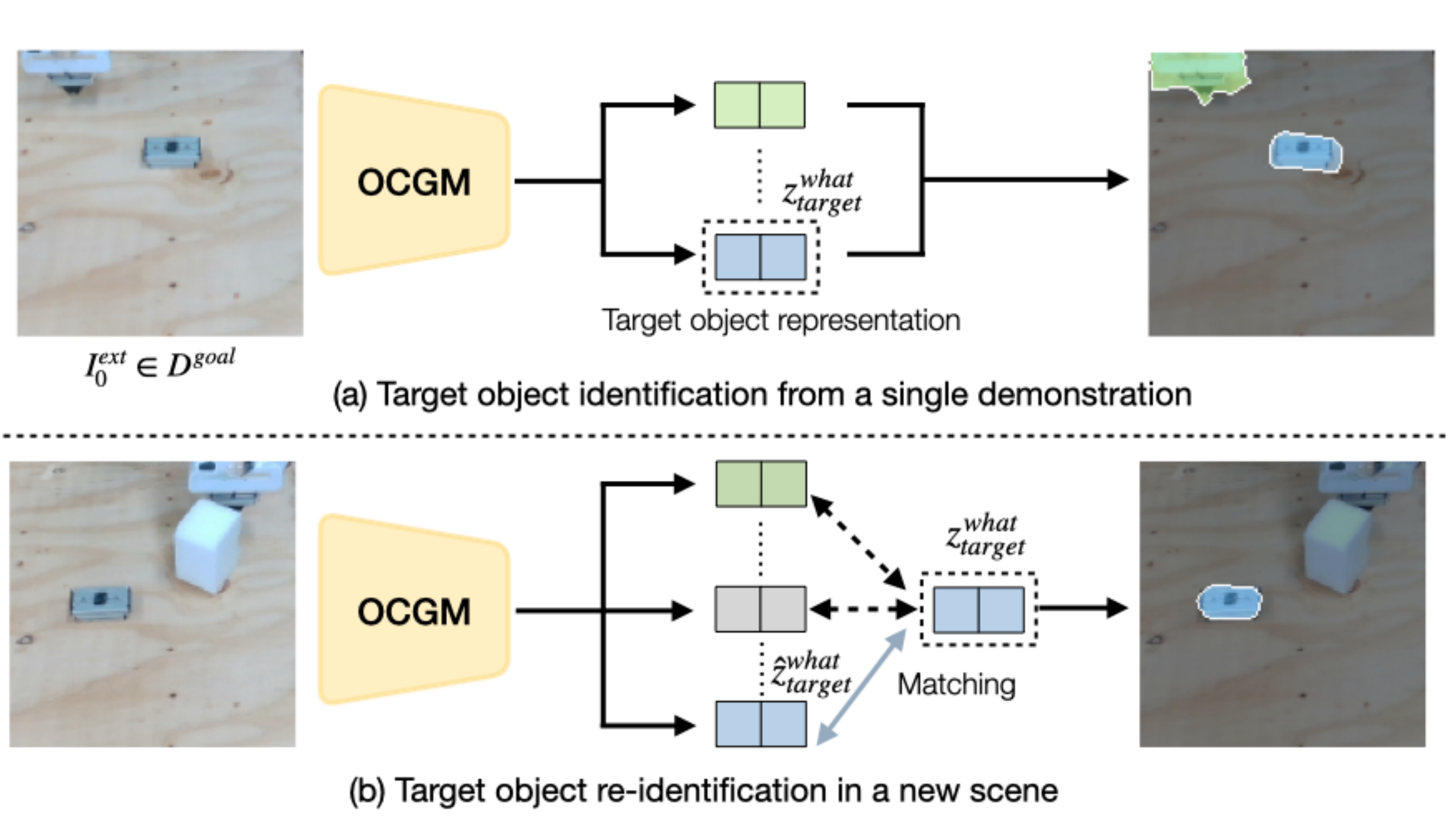}
\caption{\textbf{Target object identification and re-identification using an OCGM.} (a) We leverage a pre-trained OCGM to extract an object-centric representation from a single task demonstration. The target object is identified in the demonstration as the object mask closest to the robot end-effector position at the end of the trajectory $\boldsymbol{x}^{ee}_{T}$ (see Eq~\ref{eq:target_identification}). (b) Given a new scene, the OCGM is used to acquire object-centric representations of all objects and compare these with the already identified target object representation to re-identify the target object.}
\vspace{-0.5em}
\label{fig:target_identification}

\end{figure} 

\paragraph{Goal Identification via OCGM}
As MP requires a goal pose to plan a collision-free path through the scene, we leverage the pretrained OCGM to identify the target object from the single demonstration $D^{goal}$ and re-identify it in the current scene to specify the goal for MP.
To identify the target object from $D^{goal}$, we first acquire a set of object-centric representations by encoding the first external camera image $\boldsymbol{I}^{ext}_{0}$ in the demonstration $D^{goal}$ (see Figure~\ref{fig:target_identification} (a)).
We determine the target object-centric representation $\boldsymbol{z}^{what}_{target}$ such that the object is present at the beginning of the trajectory, \ie $p(z^{pre}_{0, j}) \geq 0.5$ and such that it is the closest to the robot end-effector position $\boldsymbol{x}^{ee}_{T}$ at the end of the demonstration $D^{goal}$.
To calculate the 3D poses of objects in the robot's reference frame, the centre of the object mask predicted by the OCGM is converted to Cartesian coordinates using the RGB-D camera's depth plane and the known camera extrinsics.
The closest object to the robot end-effector position at the end of the demonstration $D^{goal}$  is calculated using $L_{2}$ distance: 
\begin{equation}
    target = \argmin_{j, j=1..N} ||\boldsymbol{x}^{ee}_{T}-\boldsymbol{o}_{j}||_{2}
\label{eq:target_identification}
\end{equation}
where $o_{j}$ is the 3D object positions in the robot reference frame and $N$ is the number of objects discovered by the OCGM in the scene $\boldsymbol{I}^{ext}_{0}$. In order to re-identify the target object in the current scene (see Figure~\ref{fig:target_identification} (b)), we compare the target object-centric representation  $\boldsymbol{z}^{what}_{target}$ with each object-centric representation $\boldsymbol{\hat{z}}^{what}_{j}$ discovered in the new scene (see Figure~\ref{fig:target_identification} (b)) using the $L_{2}$ distance and choose the object that has the most similar representation:
 \begin{equation}
    \hat{\boldsymbol{z}}^{what}_{target} = \argmin_{j=1..N} ||\boldsymbol{z}^{what}_{target}-\hat{\boldsymbol{z}}^{what}_{j}||_{2}
\end{equation}

\paragraph{MP + Transition Policy + RL Policy}
Using the target object's pose $\boldsymbol{o}_{target}$ in the robot's reference frame, we use an RRT-connect motion planner to guide the robot's end-effector to the location of the target object (see Figure~\ref{fig:method} (b)).
To avoid collisions during the MP phase, an occupancy map, OctoMap~\cite{hornung13auro}, is created using the point clouds captured by the calibrated external camera.
After the execution of MP, we leverage the trained skill transition network to guide the arm into the initiation set of the skill (see Figure~\ref{fig:method}(c)) to maximise the outcomes of the RL policy.
Finally, the learned RL policy completes the manipulation task.

\section{Experiments}
Our experiments are designed to answer the following guiding questions: (1)~does the use of an OCGM achieve versatile and efficient target object identification for MP in the real world? (2)~how well does our system perform complex manipulation tasks, specifically industrial insertion tasks, in obstructed environments? (3)~does a skill transition network increase task success rate?

\begin{figure}[t]
    \centering
    \includegraphics[width=0.5\textwidth]{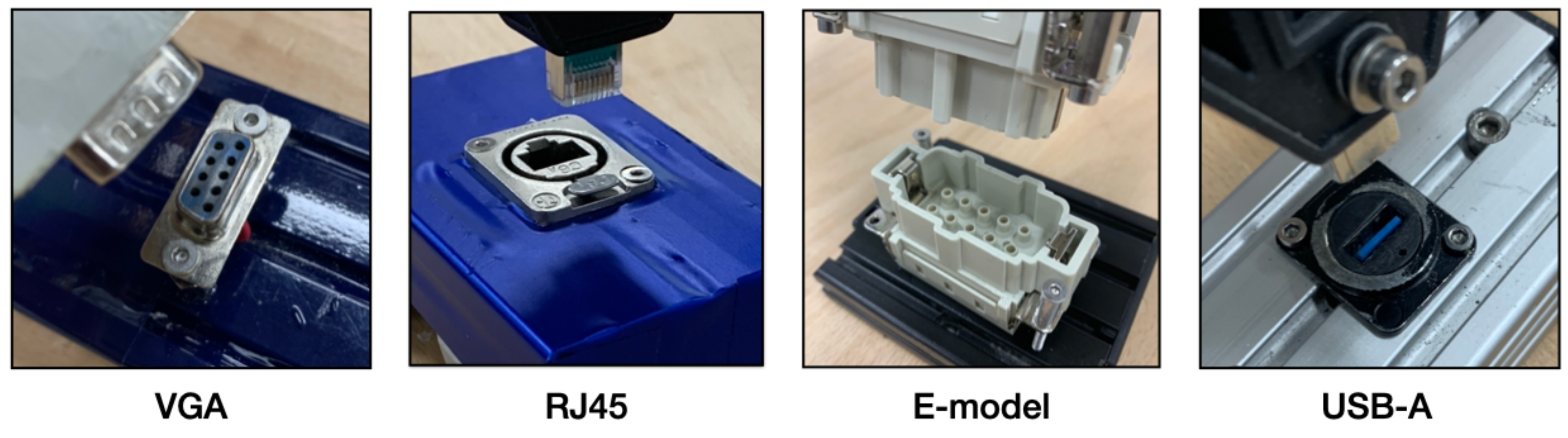}
    \caption{\textbf{Insertion tasks.} We evaluate our framework on four assembly tasks found in industry. Each socket is attached to a mount of varying size and colour to demonstrate the versatility and efficiency of our one-shot goal specification using an OCGM.}
    \label{fig:tasks}
    \vspace{-0.5cm}
    \end{figure}

\begin{table*}[t]
    \centering
    \setlength\tabcolsep{3pt}
    \scalebox{1.05}{    
    \begin{tabular}{ l  c c    c  c   c c  c c  c  c } 
     \hline
      & \multicolumn{2}{c}{VGA} & \multicolumn{2}{c}{RJ45} & \multicolumn{2}{c}{E-model} & \multicolumn{2}{c}{USB-A}  & \#Data &  Intervention\\ 
     \cmidrule(lr){2-3}\cmidrule(lr){4-5} \cmidrule(lr){6-7} \cmidrule(lr){8-9}
     Method & Accuracy & WSI & Accuracy & WSI  & Accuracy & WSI & Accuracy & WSI & & \\
     \hline
    Template matching & 70.0\% & 54.7/81.9\% & 35.0\% & 22.1/50.5\% & 87.5\% & 73.9/94.5\% & 55.0\% & 39.8/69.3\% & 1 & yes \\ 
    Feature-based matching & 40.0\% & 26.3/55.4\% & 87.5\% & 73.9/94.5\% & 7.5\% & 2.6/19.9\% & 77.5\% & 62.5/87.7\% & 1 & yes \\ 
     Object-specific classifier & 80.0\% & 65.2/89.5\% & \textbf{100.0\%} & 91.2/100.0\% & 87.5\% & 73.9/94.5\% & 75.0\% & 59.8/85.8\% & 2.5K & yes \\ 
     \hline
     OCGM identifier (Ours) & \textbf{82.5\%} & 68.0/91.3\% & 95.0\% & 83.5/98.6\% & \textbf{95.0\%} & 83.5/98.6\% & \textbf{92.5\%} & 80.1/97.4\%  & 1 & no \\ 
     \hline
    \end{tabular}
    }
    \vspace{0.15cm}
    \caption{\textbf{Accuracy of target object identification.} We evaluate our method and two baselines on 160 test scenes (40 scenes per connector), and report the accuracy, Lower Limit (LL) and Upper Limit (UL) of the Wilson score interval (WSI) with confidence interval of $95\%$. While template matching and object-specific classification requires human intervention, such as cropping a reference target object image and labelling training data, our OCGM identifier successfully identifies the target object from only a single demonstration without such human intervention.}
    \label{table:goal_identification}
\end{table*}

\subsection{Experimental Setup} 
Several industrial insertion tasks, inspired by the NIST assembly boards \cite{kenneth2020nist}, are used within our experiments (see Figure~\ref{fig:tasks}). 
To verify the robustness of the target object identification using an OCGM, mounts for each socket and arbitrary obstacles with different colour and size are used.
In our experiments, we use a Franka Panda robot (7-DOF robot arm) and rigidly attach each connector to the robot's end-effector.
Each phase of our framework uses different controllers: a joint position controller to follow a trajectory planned by the motion planner, a Cartesian pose controller for the skill transition network, and a Cartesian velocity impedance controller for the RL policy.
For each evaluation trial, the socket, robot arm, and one or two obstacles are randomly placed in the robot's operational space.
While the rotation around the z-axis of the sockets is consistent across all of the evaluation trials, we can readily extend our system to accommodate such cases by training a transition network that also predicts the displacement of the z-axis orientation to move the end-effector to the initiation set of a low-level manipulation policy. We leave this extension for future work.

Given the pre-trained OCGM, for each new manipulation skill, our modular framework requires a total of $10$ minutes of human-supervised demonstrations and a maximum of $130$ minutes of unsupervised training comprising of: up to $90$ minutes for RL policy training and $40$ minutes for data collection and training of the skill transition network.

\subsection{Efficient and Versatile Target Object Identification}
First, the OCGM identifier is evaluated against several baselines on 160 test scenes (40 for each of the four sockets) with the target locations hand-labelled with bounding boxes for quantitative comparison.
During testing, if the intersection of union (IoU) between a ground truth and the returned bounding box from the tested algorithm is greater than $0.5$, we count it as successful~\cite{Everingham15}.

We evaluate our proposed goal identification approach against three baselines. \textit{Template matching} finds the target object in the current scene by calculating a correlation coefficient given a manually cropped target object reference image. 
\textit{Feature-based matching} finds a pair of the best matched keypoints between the manually cropped target object reference image and the current scene using FLANN-based matching~\cite{flann2011} and SIFT descriptor~\cite{Lowe:2004:DIF:993451.996342}.
\textit{Object-specific classifier} trained on a dataset of manually cropped object images with binary labels, is queried with cropped images found using a region proposal method \cite{Uijlings13}. Lastly, we evaluate our method by retrieving the minimum bounding box of the target object mask predicted by the OCGM.

\paragraph{Results.}
We report the accuracy of target object identification in Table~\ref{table:goal_identification}.
Our method achieves commensurate or better performance compared to other baselines, whilst not requiring human intervention or an object-specific dataset which requires laborious manual data labelling.
This result motivates the use of OCGMs for efficient goal acquisition for MP.
Template matching often struggles to find a target object with high confidence, potentially due to the projection of a slanted camera and small objects.
Feature-based matching also shows a lower success rate for several objects due to a lack of distinguishing features, especially for small objects in a scene. 
Object-specific classifier, on the other hand, generally performs well because it is tailored to a single object, and can be further improved by collecting more data.
However, such classifiers requires re-training on a new dataset manually labelled for each new object, limiting the versatility and efficiency of goal specification.

\subsection{Insertion Tasks in Obstructed Environments}
\begin{figure*}[t!]
\centering
\includegraphics[width=0.9\textwidth]{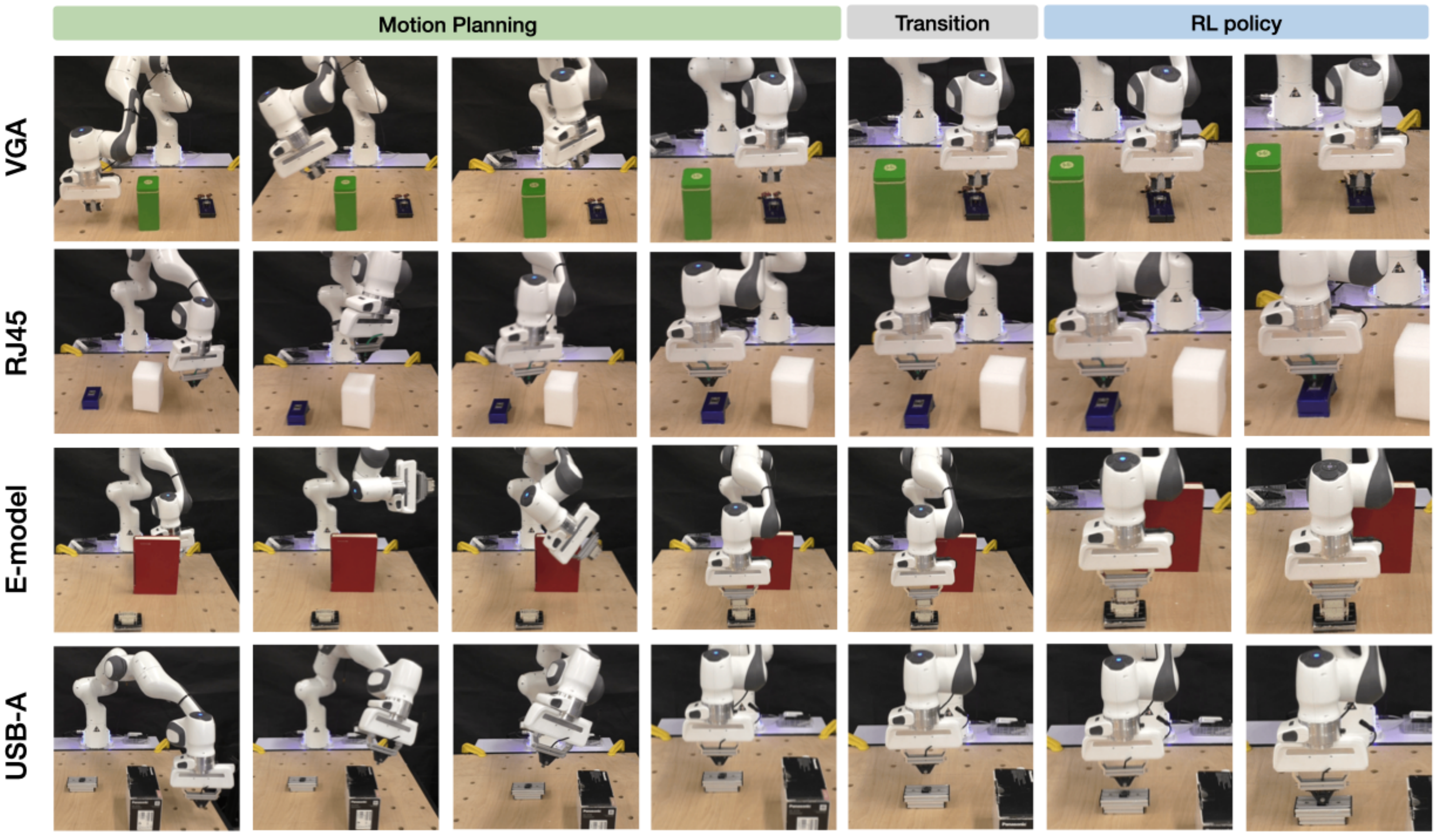}
\caption{\textbf{Real-world industrial assembly tasks in obstructed environments.} The OCGM is used to specify a goal for MP, followed by a skill transition network and a learned RL policy. Our method successfully solves complex manipulation tasks with a high success rate.}
\label{fig:task_execution}
\end{figure*}

We evaluate our proposed system on several industrial insertion tasks in obstructed environments against a series of baselines composed of competing methods. All baselines that utilise MP make use of the OCGM for target object identification. For each task, we conduct 30 trials and report the success rate in Table~\ref{table:success_results}. Figure~\ref{fig:task_execution} illustrates the execution of our method for each insertion task in the obstructed environments.

We compare the performance of our approach against a state-of-the-art RL algorithm and four comparable instantiations of our approach. \emph{Soft Actor-Critic (SAC)} a state-of-the-art RL algorithm that predicts the desired Cartesian velocity from sparse rewards, trained with 25 demonstrations using FERM~\cite{zhan2020framework} (see Appendix~\ref{appendix:sac_baseline} for training details).
\emph{MP+Demonstration Replay} substitutes replaying a single expert demonstration for the learned RL policy execution in our method, inspired by previous work~\cite{johns2021coarse-to-fine}.
\emph{MP+BC} replaces the learned RL policy in our method with Behaviour Cloning (BC)~\cite{zhang2018deep, bojarski2016end}, trained from $25$ demonstrations.
\emph{MP+Heuristic} uses a manually designed heuristic policy~\cite{luo2021shield} instead of the learned RL policy in our method to solve the task (see Appendix~\ref{appendix:heuristic_baseline} for details of the heuristic policy). 
Lastly, we evaluate our method without a skill transition network (\emph{MP+RL w/o skill transition}) for comparison.

\paragraph{Results.} 
As described in Table~\ref{table:success_results}, our method (MP+RL) as outlined in Section~\ref{sec:method} records the highest success rate for all tasks. The results for the \textit{SAC} baseline show that it is unable to solve any of the tasks, likely because it requires a large number of samples to train the policies in the robot's operational space with obstructions. \textit{MP+Demonstration Replay} is the most data-efficient method, however, it mostly fails to solve any of the tasks because it requires very accurate estimation of pose offsets for the demonstration replay to be successful.
\textit{MP+BC} is another efficient skill acquisition method because it does not require any additional interactions with the environments other than the given demonstrations to learn manipulation skills. However, due to the narrow state coverage, it struggles to solve the tasks.
While \textit{MP+Heuristic} is able to solve some insertion tasks, such as USB-A and E-model, almost one-third of the time, it fails to solve the tasks the majority of the time due to the need for accurate pose offset (the same reason for failure as \textit{MP+Demonstration Replay}).
While our method achieves high success rate over 4 industrial insertion tasks, the main failure case is caused by the misidentification of the target object by the OCGM.
These failure modes can be readily eliminated by extended and/or augmenting the OCGM training.

\begin{table*}[ht]
    \centering
    \setlength\tabcolsep{4pt}
    \scalebox{1.2}{
    \begin{tabular}{l c c  c c  c c  c c }
    \hline
      & \multicolumn{2}{c}{VGA} & \multicolumn{2}{c}{RJ45} & \multicolumn{2}{c}{E-model} & \multicolumn{2}{c}{USB-A}  \\
      \cmidrule(lr){2-3}\cmidrule(lr){4-5} \cmidrule(lr){6-7} \cmidrule(lr){8-9}
      Method & Success & WSI & Success & WSI & Success & WSI & Success & WSI  \\
     \hline
     SAC & 0.0\% & 0.0/16.1\%  & 0.0\% & 0.0/16.1\% & 0.0\% & 0.0/16.1\% & 0.0\% & 0.0/16.1\%  \\
     MP + Demonstration Replay & 3.3\% & 1.0/16.7\% & 0.0\% & 0.0\%/16.1\% & 3.3\% & 1.0/16.7\% & 0.0\% & 0.0/16.1\% \\ 
     MP + BC & 16.7\% & 7.3\%/33.6\% & 16.7 & 7.3/33.6\% & 23.3\% & 11.8/40.9\% & 26.7\% & 14.2/44.5\%  \\ 
     MP + Heuristic & 10.0\% & 3.5/25.6\% & 16.7 & 7.3/33.6\% & 36.7\% & 21.9/54.5\% & 43.3\% & 27.4/60.8\% \\ 
     \hline
     MP + RL w/o skill transition & 73.3\% & 55.6/85.8\% & 46.7\% & 30.2/63.8\% & 80.0\% & 62.7/90.5\%  & 70.0\% & 52.1/83.3\%  \\      
     MP     + RL (our method) & \textbf{86.7\%} & 70.3/94.7\% & \textbf{83.3\%} & 66.4/92.7\% & \textbf{93.3\%} & 78.7/98.2\% & \textbf{96.7\%} & 83.3/99.4\%  \\
     \hline
    \end{tabular}
    }
    \vspace{0.1cm}
    \caption{\textbf{Real-world assembly results.} We report the success rate, Lower Limit (LL) and Upper Limit (UL) of the Wilson score interval (WSI) with confidence interval of $95\%$ over 30 trials. Our method outperforms, by a significant margin, all of the other methods including a state-of-the-art RL method and several comparable instantiations of our method.}
    \label{table:success_results}

\vspace{-1em}
\end{table*}

Examining whether the transition network is required to solve complex manipulation tasks in obstructed environments (see Table~\ref{table:success_results}), the results verify that using the skill transition network results in higher success rates than without the skill transition policy.
Due to errors caused by the OCGM, camera extrinsics and estimation of the 3D goal poses, a terminal state of MP can often be outside of the initiation set of the learned RL skill.
Therefore, by introducing the skill transition module to move the robot arm into the initiation set of the skill, we can mitigate these issues and achieve better performance.

\section{Conclusion}

In this work, we propose a modular system that leverages an OCGM for one-shot goal identification and re-identification as a vital component to combine MP and RL to solve complex manipulation tasks in obstructed environments.
Specifically, the OCGM extracts a target object from only a single demonstration and re-identifies the object to determine a goal pose for MP without the need of fine-tuning on an object-specific dataset.
The experimental results show that our method for goal specification using an OCGM achieves better versatility and comparable accuracy to other tested baselines. In addition, our method successfully solves real-world industrial insertion tasks in obstructed environments from few demonstrations.
In future work we will look to investigate more advanced settings, such as randomising the socket orientation.

\section*{ACKNOWLEDGMENT}
This work was supported by a UKRI/EPSRC Programme Grant [EP/V000748/1], we would also like to thank the University of Oxford for providing Advanced Research Computing (ARC) (http://dx.doi.org/10.5281/zenodo.22558) and the SCAN facility in carrying out this work.

\bibliographystyle{IEEEtran}
\bibliography{bib/conference,bib/deep_learning,bib/rl,bib/env,bib/robotics, bib/motion_planning, bib/ocgm, bib/il, bib/misc}

\appendix
\section{Pre-training Details}
\label{appendix:pretraining}
\subsection{APEX Training Details and Hyperparameters}
\label{appendix:apex}
We collect a synthetic dataset consisting of $30K$ trajectories each of which has $10$ to $15$ frames simulated using RLBench~\cite{james2020rlbench}.
In each trajectory, a robot randomly pushes objects on a table.
To apply the pre-trained APEX to the real world, we add a small amount of noise, ranging from $1$cm to $-1$cm, to the Cartesion camera position in the simulation so that the background texture changes slightly for each trajectory and prevents APEX from overfitting to the specific background texture.
Table~\ref{table:apex_hyper} shows hyperparameters of APEX.
For further details, see prior work~\cite{wu2021Apex}.

\begin{table}[h]
\centering
\caption{APEX hyperparameter}
\begin{tabular}{|c|c|} 
 \toprule
 Parameter & Value  \\ 
 \midrule
 Observation Rendering & (128, 128), RGB \\
 Optimizer & Adam  \\ 
 Learning rate & 1e-4\\
 horizon & 10 \\
 Image size & 128 \\
 Foreground std & 0.11 \\
 Background std & 0.04 \\
 KL divergence for $z_{what}$ & 3e-4 \\
 KL divergence for $z_{where}$ & 15 \\
 KL divergence for $z_{pre}$ (discovery) & 32 \\
 KL divergence for $z_{pre}$ (tracking) & 1 \\
 Background reconstruction loss weight & 10 \\
 foreground reconstruction loss weight & 1 \\
 \bottomrule
\end{tabular}
\label{table:apex_hyper}
\end{table}

\section{In-Situ Training Details}
\label{appendix:in_situ}

\subsection{RL Training Details and Hyperparameters}
\label{appendix:RL}
We define a sparse reward function such that a policy receives $1$ when the L2 distance between robot's end-effector pose and the pre-defined target pose is within a small tolerance of $0.8$cm for VGA and $1$cm for the other insertion tasks. 
The policy and critic take as input a wrist camera image, Cartesian velocity, and F/T sensor values of the end-effector. The critic also takes as input an action sampled from the policy.
The architecture of the policy and the critic includes a CNN for processing the wrist camera image followed by a concatenation with state information. The concatenated features are fed to the policy and critic feedforward neural networks with the outputs being actions and values respectively. 
Similar to FERM~\cite{zhan2020framework}, we apply random crops and brightness changes to the image observations to acquire a robust policy efficiently.
The policy runs at $10$Hz and we find that this is sufficient for completing the complex insertion tasks.
Table~\ref{table:ferm} shows hyperparameters.

\begin{table}[h]
\centering
\caption{FERM hyperparameter}
\begin{tabular}{|c|c|} 
 \toprule
 Parameter & Value  \\ 
 \midrule
 Optimizer & Adam  \\ 
 Learning rate & 1e-3\\
 Discount factor ($\gamma$) & 0.99 \\
 Replay buffer size & $10^4$ \\
 Latent dimension & 50 \\
 Convolution filters & $[8, 16, 32, 64]$ \\
 Convolution strides & $[2, 2, 2, 2]$ \\
 Convolution filter size & 3 \\
 Hidden Units (MLP) & [128] \\
 Nonlinearity & Leaky ReLU \\
 Target smoothing coefficient ($\tau$) & 0.005 \\
 Target update interval & 2 \\
 Actor update interval & 2 \\
 Network update per environment step & 10 \\
 \bottomrule
\end{tabular}
\label{table:ferm}
\end{table}

\subsection{Transition Network Details and Hyperparameters}
\label{appendix:transition}
The dataset for the transition network is collected by uniformally sampling a Cartesian offset from within $\pm5$cm of the initial position used for RL training.
The robot moves linearly between the initial position used for RL training and the sampled offset, recording both the wrist camera image and the current offset at $15$ steps throughout. After collecting $100$ of these trajectories, in less than $30$ minutes, the skill transition network can be trained.

\subsection{Baseline Details - SAC}
\label{appendix:sac_baseline}
\begin{figure}
    \centering
    \includegraphics[width=0.6\linewidth]{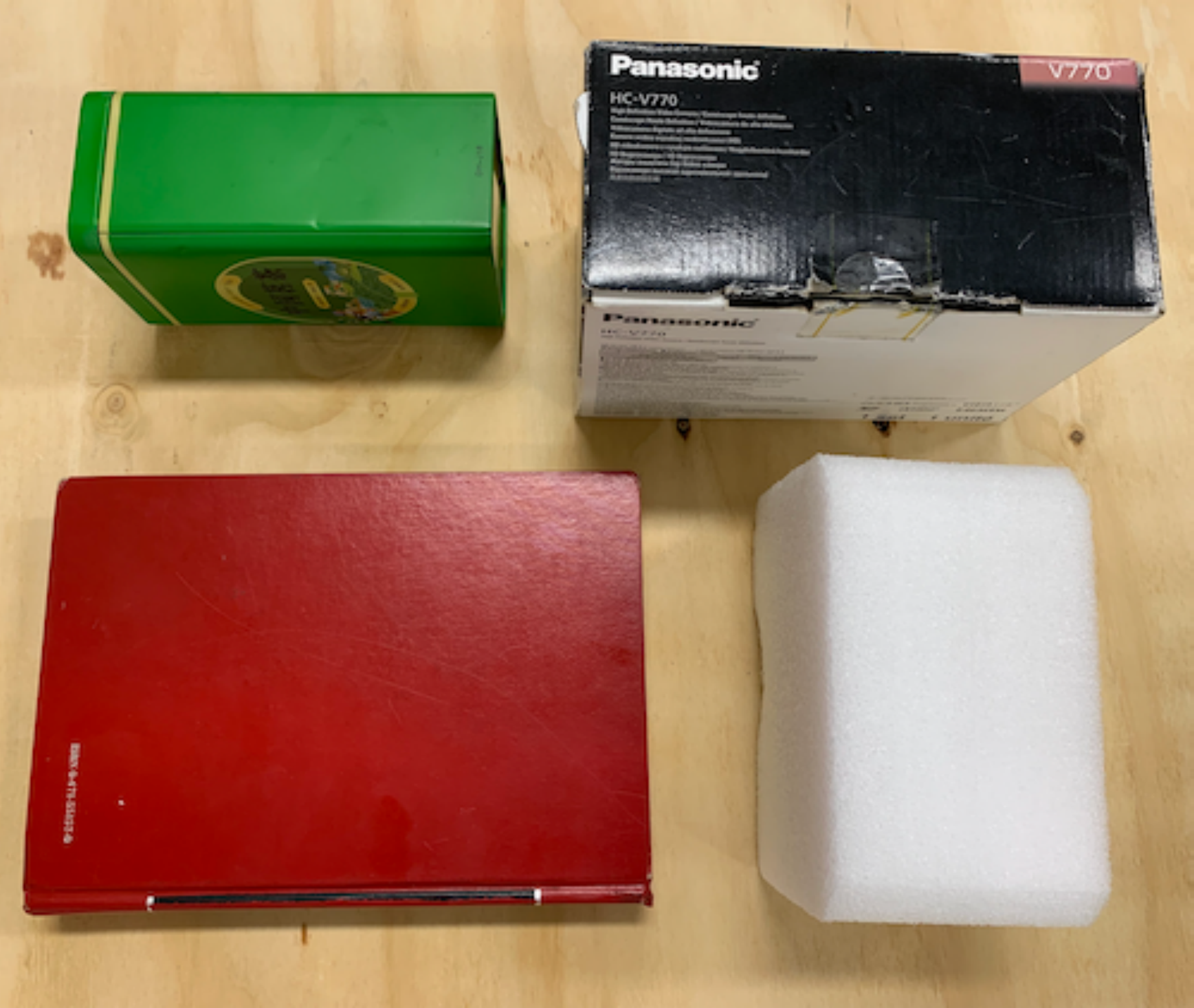}
    \caption{
        \textbf{The set of obstacles used in our experiments.} In the evaluation, we randomly sample one or two obstacles from this set of obstacles that vary in size and colour.
    }
    \label{fig:obstacle}
\end{figure}

To train a policy with SAC~\cite{haarnoja2018sac} from scratch using FERM~\cite{zhan2020framework}, we randomly place a socket and one or two obstacles from our set of obstacles (see Figure~\ref{fig:obstacle}) on a table. Then, we define a reward function for collision avoidance in addition to a sparse reward for solving the insertion task within the entire operational space of the robot:

\begin{equation}
    R = I[s \in S_{g}] - 0.005 \cdot I[s\in S_{coll}]
\end{equation}
where $S_{g}$ and $S_{coll}$ are a set of goal states and states involving collisions and $I$ is an indicator function. Furthermore, the policy also takes as input an image from the external camera in addition to a wrist camera image. We use the same hyperparameters described in Table~\ref{table:ferm}.

\subsection{Baseline Details - MP + Heuristic}
\label{appendix:heuristic_baseline}
We design a heuristic policy for insertion tasks, inspired by previous work~\cite{luo2021shield} that utilises the F/T snesor data.
The heuristic policy first moves the end-effector downwards until the arm contacts an object. Then, the arm follows an outwards spiral pattern on a 2D grid to find the socket hole. Finally, after the connector is aligned with the hole, the arm moves downwards again with a small circular movement in the X-Y plane to insert the connector into the socket.

\end{document}